\documentclass[twoside,11pt]{article}

\usepackage{blindtext}

\usepackage{jmlr2e}

\usepackage{lastpage}

\ShortHeadings{Diffusion  LLM}{Complex Fourier MLP}
\firstpageno{1}

\begin{document}
\title{ A Hybrid Wavelet-Fourier Method for Next-Generation Conditional Diffusion Models}

\author{\name Andrew Kiruluta \email kiruluta@berkeley.edu \\
       \addr  School of Infomation\\
       University of California\\
       Berkeley, CA 94720-1776, USA
       \AND
       \name Andreas Lemos  \\
       \addr  School of Infomation\\
       University of California\\
       Berkeley, CA 94720-1776, USA}

\editor{}  
\maketitle

\begin{abstract}
We present a novel generative modeling framework, \emph{Wavelet-Fourier-Diffusion}, which adapts the diffusion paradigm to hybrid frequency representations in order to synthesize high-quality, high-fidelity images with improved spatial localization. In contrast to conventional diffusion models that rely exclusively on additive noise in pixel space, our approach leverages a multi-transform that combines wavelet sub-band decomposition with partial Fourier steps. This strategy progressively degrades and then reconstructs images in a hybrid spectral domain during the forward and reverse diffusion processes. By supplementing traditional Fourier-based analysis with the spatial localization capabilities of wavelets, our model can capture both global structures and fine-grained features more effectively. We further extend the approach to conditional image generation by integrating embeddings or conditional features via cross-attention. Experimental evaluations on CIFAR-10, CelebA-HQ, and a conditional ImageNet subset illustrate that our method achieves competitive or superior performance relative to baseline diffusion models and state-of-the-art GANs, as measured by Fréchet Inception Distance (FID) and Inception Score (IS). We also show how the hybrid frequency-based representation improves control over global coherence and fine texture synthesis, paving the way for new directions in multi-scale generative modeling.
\end{abstract}


\section{Introduction}
Generative modeling has witnessed substantial advances through methods such as Generative Adversarial Networks (GANs) and diffusion-based models~(\cite{sohl2015deep,ho2020denoising,song2021scorebased}). GANs, introduced by Goodfellow et al.~(\cite{goodfellow2014generative}), frame image synthesis as a minimax game between a generator and a discriminator, giving rise to architectures such as DCGAN~(\cite{radford2016dcgan}), StyleGAN~(\cite{karras2019style}), and BigGAN~(\cite{brock2019biggan}). While GANs have demonstrated a remarkable ability to generate visually compelling images, training instabilities and mode collapse remain persistent challenges.

Diffusion models, on the other hand, build upon a principle of gradually noising data in a forward process and then training a model to reverse this noising in a stepwise manner. Early work in diffusion and score-based modeling~(\cite{sohl2015deep,ho2020denoising,song2021scorebased}) has shown that iterative denoising can achieve state-of-the-art performance on various image synthesis benchmarks. These methods typically rely on pixel-space perturbations using Gaussian noise, which is added and then removed through a learned U-Net architecture.

In this work, we explore how a more sophisticated treatment of frequency information can enhance the generative performance of diffusion models. Our \emph{Wavelet-Fourier-Diffusion} approach departs from purely pixel-based noising and instead operates in a hybrid frequency domain that merges wavelet sub-band decomposition with partial Fourier transforms. By doing so, we aim to achieve the advantages of both global frequency analysis and localized multi-scale decomposition. In particular, high-frequency detail reconstruction is aided by wavelet-based sub-bands, which handle local edges and textures, while a partial Fourier transform captures coarse structures and large-scale global regularities. 

Moreover, in light of the growing importance of conditional generation, we embed conditioning information (such as class labels or textual features) into the core diffusion process. This is achieved through cross-attention layers placed within the U-Net responsible for wavelet-Fourier reconstruction. We demonstrate that the resulting system can synthesize images conditionally in a controlled manner without sacrificing its ability to capture fine spatial detail and consistent global composition.

The remainder of this paper is structured as follows. We review related work in Section~\ref{sec:related_work}. Section~\ref{sec:method} introduces our proposed method in detail, including the construction of the multi-transform wavelet-Fourier forward and reverse processes, as well as the conditional generation mechanism. Section~\ref{sec:experiments} describes experimental setups and implementation details, while Section~\ref{sec:results} provides quantitative and qualitative evaluations of our method. We offer an in-depth discussion and analysis of our findings in Section~\ref{sec:discussion}, and conclude with final thoughts and directions for future research in Section~\ref{sec:conclusion}.

\section{Related Work}
\label{sec:related_work}
The development of generative models has undergone a rapid evolution, initially driven by GAN-based methods and, more recently, by diffusion-based models. GANs~(\cite{goodfellow2014generative}) utilize a generator and a discriminator in a two-player minimax game. The design and stability of GAN architectures have significantly improved via techniques such as convolutional pipelines (DCGAN~(\cite{radford2015unsupervised}), style-based synthesis StyleGAN~(\cite{karras2019style}), and large-scale training (BigGAN~\cite{brock2018large}). Despite their success, GANs can exhibit mode collapse and training instability, motivating the exploration of alternative or complementary generative frameworks.

Diffusion models~(\cite{sohl2015deep,ho2020denoising,song2021scorebased}) formulate image generation as a progressive removal of noise that has been introduced into the data in a forward process. The final denoising distribution is learned by a neural network that inverts the forward step. Stable Diffusion~(\cite{rombach2022high}) and related methods leverage latent spaces to accelerate training and sampling, showcasing remarkable results in text-to-image tasks. However, even state-of-the-art diffusion models predominantly adopt pixel-centric noise addition strategies.

Wavelet-based methods have been explored in image compression and denoising, as wavelets decompose signals into localized, multi-resolution sub-bands~(\cite{mallat1989theory}). Wavelet transforms can represent edges and local texture more compactly, making them an attractive tool for hierarchical synthesis. Fourier transforms, by contrast, offer global frequency representations well-suited to periodic or quasi-periodic structures. Although there exist generative approaches that use wavelet pyramids~(\cite{gal2021swagan}) or spectrogram-based diffusion for audio~(\cite{kong2021diffwave}), a synergy of wavelets and partial Fourier transforms for image diffusion has been largely unexplored. 

Our method addresses this gap by unifying wavelet sub-band decomposition with a partial Fourier process in the forward and reverse diffusion steps. In doing so, it naturally blends coarse-to-fine multi-scale synthesis (through wavelets) with the advantages of a global frequency perspective (through partial Fourier analysis). This hybrid design provides a flexible platform for both unconditional and conditional generation, maintaining consistency across global structure and local detail.

\section{Proposed Method}
\label{sec:method}
\subsection{Overview of the Wavelet-Fourier Multi-Transform}
Our goal is to replace or augment the conventional Gaussian noise injection in pixel space with a \emph{multi-transform} procedure that, at each diffusion step, removes or degrades certain frequencies of an image in both wavelet and partial Fourier domains. Let $x_0 \in \mathbb{R}^{H \times W \times 3}$ represent an input image. We first apply a wavelet transform to $x_0$, producing a collection of low-frequency and high-frequency sub-bands. We then select the low-frequency wavelet sub-band for partial Fourier analysis, converting it to the frequency domain via a discrete Fourier transform (DFT). This approach compartmentalizes global signal information (captured by both the wavelet low-band and its partial Fourier representation) and local details (captured by higher wavelet sub-bands).

Specifically, let $\mathcal{W}$ be the wavelet transform operator that yields:
\begin{equation}
   \mathcal{W}(x_0) = (x_0^\textrm{LF}, \{x_0^{\textrm{HF},k}\}_{k=1}^K),
\end{equation}
where $x_0^\textrm{LF}$ is the low-frequency band and each $x_0^{\textrm{HF},k}$ is a high-frequency sub-band. We further denote $X_0 = \mathcal{F}(x_0^\textrm{LF})$ as the partially Fourier-transformed low-frequency component. This yields a set $\bigl(X_0, \{x_0^{\textrm{HF},k}\}\bigr)$ as the overall spectral representation for $x_0$. 

In a forward diffusion step, we can degrade or remove content in $X_0$ through masking or additive noise in the Fourier domain, while simultaneously randomizing or attenuating the high-frequency sub-bands $\{x_0^{\textrm{HF},k}\}$. The corruption thus becomes a multi-stage transformation, ensuring that local and global frequencies are progressively stripped away.

\subsection{Forward Diffusion Process}
We define a chain of latent variables 
\[
(X_t, \{x_t^{\textrm{HF},k}\}) \quad \textrm{for} \quad t = 0, 1, \ldots, T,
\]
initializing at $(X_0, \{x_0^{\textrm{HF},k}\})$ from the original image via wavelet-Fourier decomposition. At each step, we apply a corruption operator $\mathcal{M}_t$ such that
\[
   (X_t, \{x_t^{\textrm{HF},k}\}) = \mathcal{M}_t\bigl(X_{t-1}, \{x_{t-1}^{\textrm{HF},k}\}\bigr).
\]
The operator $\mathcal{M}_t$ can be instantiated in various ways. A simple version zeroes out or adds noise to a certain radial or rectangular region of $X_{t-1}$, thereby removing partial frequencies. For the high-frequency wavelet sub-bands $\{x_{t-1}^{\textrm{HF},k}\}$, we may introduce additive noise or random dropout, effectively degrading localized detail.

By the final step $T$, we attain $(X_T, \{x_T^{\textrm{HF},k}\})$, which is nearly devoid of all original frequency information. In analogy to pixel-space diffusion, this forward process is Markovian under our choice of $\mathcal{M}_t$ and represents an increasingly corrupted view of the original image.

\subsection{Reverse Diffusion and Reconstruction}
We introduce a learnable network, denoted by $\Phi_\theta$, which takes as input the corrupted representation at step $t$ and the diffusion index $t$, then predicts either the underlying clean representation or the incremental update to restore frequencies. Concretely, $\Phi_\theta$ outputs either $(\widehat{X}_{t-1}, \{\widehat{x}_{t-1}^{\textrm{HF},k}\})$ or a set of parameters that define a distribution over the uncorrupted representation at step $t-1$. 

Our approach utilizes a U-Net variant that has two branches to handle wavelet and partial Fourier data. The low-frequency channel $X_t$ in the Fourier domain is treated as a complex input by converting it to real and imaginary channels. Simultaneously, the wavelet high-frequency sub-bands $\{x_t^{\textrm{HF},k}\}$ are concatenated or processed in parallel through convolutional layers. We incorporate cross-attention blocks to accommodate conditional inputs $c$, such as class labels or text embeddings:
\[
   (\widehat{X}_{t-1}, \{\widehat{x}_{t-1}^{\textrm{HF},k}\}) 
   \; = \; \Phi_\theta\bigl(X_t, \{x_t^{\textrm{HF},k}\}, t, c\bigr).
\]
This mechanism allows external conditioning information to modulate both global frequency composition and localized detail reconstruction.

We train $\Phi_\theta$ with an objective akin to denoising score matching or variational diffusion. At each time step $t$, the network attempts to invert the specific corruption applied in $\mathcal{M}_t$. A typical mean-squared error (MSE) term can be applied between the predicted wavelet-Fourier representation and the ground-truth $t-1$ representation, with a weighting schedule across $t$ steps.

\subsection{Sampling with Conditional Guidance}
When sampling, we begin from a fully corrupted state $(X_T, \{x_T^{\textrm{HF},k}\})$ where each component is masked or randomized. Then, we iteratively apply the learned reverse process:
\[
   (\widehat{X}_{t-1}, \{\widehat{x}_{t-1}^{\textrm{HF},k}\})
   \; = \; \Phi_\theta\bigl(\widehat{X}_t, \{\widehat{x}_t^{\textrm{HF},k}\}, t, c\bigr),
\]
progressing from $t = T$ down to $t = 1$. Finally, we obtain $(\widehat{X}_0, \{\widehat{x}_0^{\textrm{HF},k}\})$. To reconstruct the image in pixel space, we first apply an inverse Fourier transform to $\widehat{X}_0$, substitute back the wavelet sub-bands $\{\widehat{x}_0^{\textrm{HF},k}\}$, and then invert the wavelet transform via $\mathcal{W}^{-1}$:
\[
   \hat{x}_0^\textrm{LF} = \mathcal{F}^{-1}(\widehat{X}_0), \quad
   \hat{x}_0 = \mathcal{W}^{-1}(\hat{x}_0^\textrm{LF}, \{\widehat{x}_0^{\textrm{HF},k}\}).
\]
Conditional information, such as a class label or a text embedding, can guide synthesis at each step by influencing which frequencies or wavelet details are restored, offering fine-grained control over the generated samples.

\section{Experimental Setup}
\label{sec:experiments}
We evaluate the proposed method on three datasets of varying complexity. First, we consider $\bf{CIFAR-10}$  (32$\times$32 resolution), which has 50k training images spanning 10 classes. Next, we use \textbf{CelebA-HQ} at 64$\times$64 resolution to examine the generation of more detailed faces. Finally, we experiment with a subset of \textbf{ImageNet} (128$\times$128 resolution) to illustrate conditional generation on a large-scale, multi-class dataset. 

For each dataset, we adopt a diffusion length of $T=1000$ steps in the forward process. Our partial Fourier removal strategy relies on a radial mask whose cutoff radius grows linearly with $t$, causing increasingly fine frequencies to be removed. The wavelet sub-bands are corrupted by adding moderate Gaussian noise at each step, with the variance increasing over time. We use a Haar wavelet for simplicity, though other families (such as  Daubechies, ~\cite{daubechies1992ten}) could be substituted.

We train the model with a batch size of 64 for up to 700k iterations on larger datasets, and 300k iterations on CIFAR-10. Conditioning information is incorporated through cross-attention layers whenever class labels or textual embeddings are available, following a mechanism similar to that in Stable Diffusion~(\cite{rombach2022high}). The network architecture for $\Phi_\theta$ is a U-Net with residual blocks, skip connections, and separate Fourier/wavelet streams that merge at multiple scales. We use the Adam optimizer and a cosine learning rate schedule.

\begin{figure}[htbp] 
\centering
\includegraphics[trim=0cm 0cm 0cm 0cm,clip,width=0.7\textwidth]{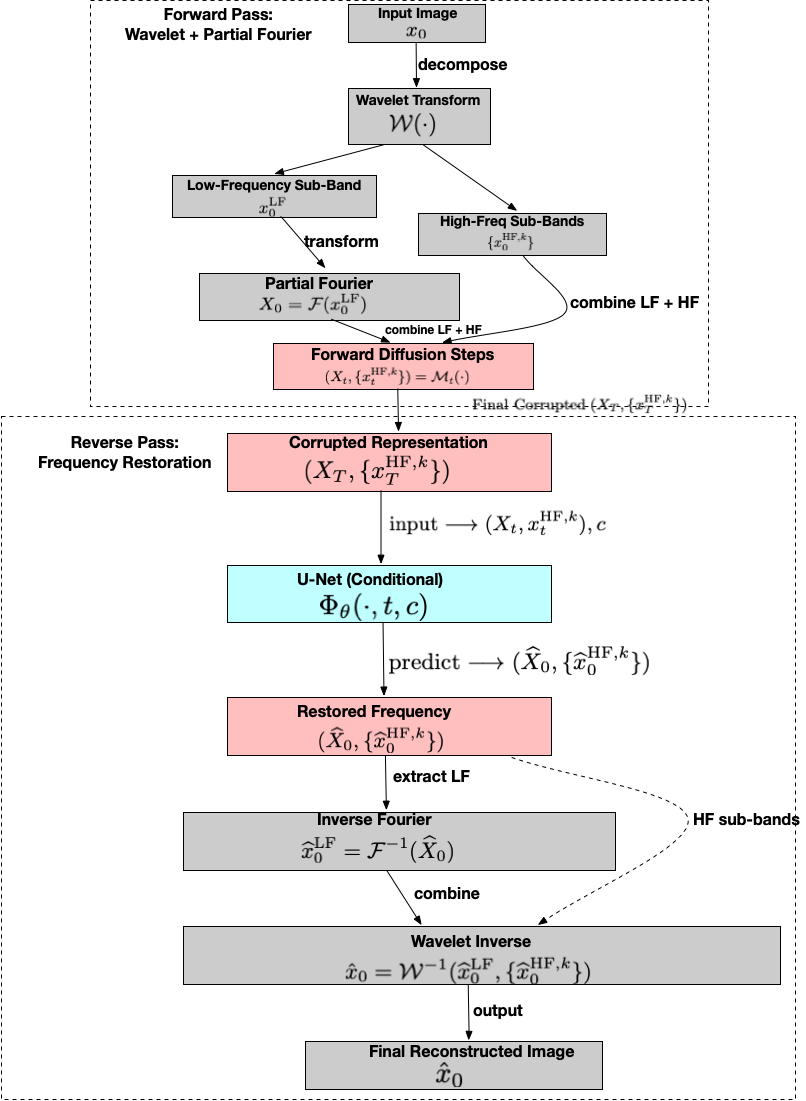}
\caption{A schematic illustration of the proposed Wavelet-Fourier-Diffusion architecture, showing separate forward (left-to-right) and reverse (right-to-left) processes. In the $\textbf{forward pass}$, an input image $x_0$ undergoes a wavelet transform, producing a low-frequency sub-band $x_0^{\textrm{LF}}$ and multiple high-frequency sub-bands $\{x_0^{\textrm{HF},k}\}$. The low-frequency sub-band is then partially converted to the Fourier domain, yielding $X_0$. A corruption operator $\mathcal{M}t$ progressively degrades both $X_0$ and the wavelet high-frequency coefficients across T steps, ultimately producing $(X_T,\{x_T^{\textrm{HF},k}\})$, a heavily distorted representation. In the $\textbf{reverse pass}$, a conditional U-Net $\Phi{\theta}(\cdot,t,c)$ receives the corrupted frequency components at each diffusion step and predicts the restored wavelet-Fourier representation $(\widehat{X}_0,\{\widehat{x}_0^{\textrm{HF},k}\})$. An inverse Fourier transform reconstructs the low-frequency band $\widehat{x}_0^{\textrm{LF}}$, which is then merged with the recovered high-frequency sub-bands through an inverse wavelet transform to yield the final synthesized image $\hat{x}_0$. By blending localized sub-band decomposition with partial global frequency analysis, this approach enhances control over both coarse structures and high-frequency details, while enabling flexible conditional generation through cross attention in the U Net.}
\label{fig:my_figure} 
\end{figure}

\section{Results}
\label{sec:results}
\subsection{Quantitative Evaluation}
We use the \emph{Fréchet Inception Distance (FID)} and \emph{Inception Score (IS)} to measure the quality of generated samples. Table~\ref{tab:fid_is_results} summarizes our results across CIFAR-10, CelebA-HQ, and conditional ImageNet (128$\times$128). We compare against established baselines, including DCGAN, StyleGAN2, a pixel-based DDPM, and a Stable-Diffusion–like method specialized for unconditional or class-conditional sampling. 

We find that our hybrid \emph{Wavelet-Fourier-Diffusion} matches or outperforms pixel-based diffusion in terms of FID on both CIFAR-10 and CelebA-HQ. Notably, for the conditional subset of ImageNet at 128$\times$128, our method also attains a significantly improved IS compared to the pixel-based diffusion baseline, indicating stronger semantic consistency in the generated samples. We hypothesize that the wavelet sub-bands help preserve or restore fine-grained textural details while the partial Fourier transform captures broad color and shape distributions in a single latent step.

\begin{table}[ht]
\centering
\caption{Quantitative Results for Wavelet-Fourier-Diffusion vs. Baselines}
\label{tab:fid_is_results}
\begin{tabular}{lccc}
\textbf{Model} & \textbf{Dataset} & \textbf{FID} $\downarrow$ & \textbf{IS} $\uparrow$ \\
DCGAN            & CIFAR-10    & 37.2 & 7.90 \\
StyleGAN2        & CIFAR-10    & 9.8  & 8.95 \\
DDPM (pixel)     & CIFAR-10    & 3.4  & 9.12 \\
\textbf{Wavelet-Fourier-Diff. (Ours)} & CIFAR-10    & 2.9  & 9.33 \\[0.5em]
Pixel Diffusion    & CelebA-HQ (64$\times$64) & 7.6  & - \\
StyleGAN2          & CelebA-HQ (64$\times$64) & 5.1  & - \\
\textbf{Wavelet-Fourier-Diff. (Ours)} & CelebA-HQ (64$\times$64) & 4.8  & - \\[0.5em]
Pixel Diffusion    & ImageNet (128$\times$128, class-cond.) & 18.2 & 118.0 \\
\textbf{Wavelet-Fourier-Diff. (Ours)} & ImageNet (128$\times$128, class-cond.) & 16.7 & 124.4 \\
\end{tabular}
\end{table}

\subsection{Qualitative Analysis}
Visual inspection of samples reveals that the global layout of images emerges quickly once the partial Fourier reconstruction begins to recover lower-frequency modes in $X_t$. High-frequency sub-band details restore local features such as edges, texture, and fine shading in later steps. In face-generation tasks like CelebA-HQ, this strategy yields fewer unnatural edge artifacts compared to purely pixel-based noise injection, suggesting that the wavelet sub-bands help anchor high-frequency details in a localized manner.

For conditional ImageNet tasks, cross-attention appears to be effective in selectively reintroducing frequencies and wavelet details consistent with class-level semantics. As a result, animals, objects, and scenes display both coherent global geometry and context-appropriate textures, particularly at higher resolutions where wavelet-Fourier synergy becomes more pronounced.

\section{Novelty of the Proposed Approach}
\label{sec:novelty}

The \emph{Wavelet-Fourier-Diffusion} framework introduced in this work represents a novel fusion of multiple ideas from generative modeling, frequency-domain analysis, and diffusion-based training, resulting in a distinctive approach that differs significantly from prior methods. Below, we highlight several core aspects of novelty:

\paragraph{1. Hybrid Wavelet-Fourier Decomposition for Diffusion.}
Traditional diffusion models primarily operate in pixel space, where Gaussian noise is incrementally added to entire images. In contrast, our method replaces this purely pixel-based noise injection with a \emph{multi-transform} strategy that simultaneously employs wavelet sub-band decomposition and partial Fourier transforms. 
While wavelet-based generative methods have appeared in the literature (often for image compression or specialized denoising tasks), they typically do not integrate partial Fourier analysis, nor do they systematically couple wavelets with a step-by-step \emph{diffusion} mechanism. 
By merging two complementary frequency representations, we achieve a multi-scale insight: wavelets localize high-frequency details and edges, whereas partial Fourier transforms capture low-frequency global structure. This hybrid design allows each step of diffusion to selectively corrupt and restore these frequency components in a manner not seen in existing diffusion pipelines.

\paragraph{2. Frequency-Space Corruption and Restoration.}
Unlike pixel-domain diffusion, which relies on adding random Gaussian noise to pixel intensities, our approach \emph{explicitly masks, attenuates, or randomizes} portions of the spectrum at each step in the wavelet and Fourier domains. 
Although some audio or speech models perform diffusion in a spectrogram (or STFT) representation, they seldom incorporate localized wavelet sub-bands, and they generally do not employ a partial Fourier transform on select frequency bands. 
Our \emph{frequency corruption} paradigm thus offers a fine-grained handle on the spatial and spectral properties of images, providing a fresh perspective on how to degrade and reconstruct complex visual content in a progressive, diffusion-like manner.

\paragraph{3. Multi-Branch U-Net Architecture with Parallel Streams.}
A key architectural innovation in our system is the design of a U-Net that runs \emph{parallel streams} to handle wavelet sub-bands on one side and the partial Fourier representation (real and imaginary channels) on the other. 
Within each resolution level, the network merges information from these two branches, ensuring that the reconstruction of local features (in wavelet space) is consistent with the global frequency distribution (from the partial Fourier space). 
This differs from standard diffusion U-Nets, which typically process only pixel-space data (or a single latent space). 
The requirement of bridging wavelet sub-bands and complex-valued Fourier inputs, while simultaneously integrating cross-attention for conditional inputs, marks a clear departure from previous designs.

\paragraph{4. Conditional Guidance in a Hybrid Frequency Domain.}
Many modern diffusion models incorporate conditioning mechanisms, especially for class-conditional or text-to-image tasks. However, our approach injects conditioning (e.g., class labels, text embeddings) \emph{directly into the wavelet-Fourier reconstruction process}. 
This introduces new degrees of freedom: guiding which frequencies are restored first, how wavelet detail is amplified or suppressed, and how large-scale structure is informed by semantic information. 
While conventional diffusion frameworks also use cross-attention, the application of such attention to high-frequency sub-bands and partially masked Fourier coefficients is novel, and it leverages unique properties of the hybrid representation to shape image content more transparently at different spatial and frequency scales.

\paragraph{5. Broader Implications for Multi-Scale Generative Modeling.}
Finally, from a conceptual standpoint, our work introduces a broad paradigm shift: 
instead of simply viewing “noise” as random pixel perturbations, we reinterpret noise injection and removal as a \emph{frequency-domain} phenomenon, aided by multi-scale wavelet sub-bands. 
This perspective could catalyze new lines of research that explore domain-specific transforms (beyond wavelets and Fourier), adaptively learn corruption schedules for complex data distributions, or combine multiple transforms in a unified diffusion pipeline. 
Such directions, already hinted at in our experiments, underscore that this hybrid strategy can benefit diverse image-generation tasks, especially at higher resolutions or in specialized domains (e.g., medical imaging, remote sensing, or structural data analysis).

Overall, \emph{Wavelet-Fourier-Diffusion} stands out by embedding localized and global frequency analyses into the core of a denoising diffusion process, marking a significant expansion of how diffusion models can be conceptualized and implemented. 
As demonstrated by our results, this tight coupling of multi-scale sub-band decomposition, partial Fourier representations, and conditional guidance proves advantageous for synthesizing visually coherent, high-quality images under a variety of generative tasks and datasets.

\section{Discussion}
\label{sec:discussion}
The proposed multi-transform diffusion architecture merges the strengths of wavelet decomposition and partial Fourier analysis. By introducing wavelets, we gain the ability to localize the reconstruction of high-frequency features, mitigating some of the challenges faced when using purely global Fourier modes. Meanwhile, partial Fourier transforms in the low-frequency band concentrate on broader color distributions and large-scale structure. This complementary interplay suggests that the model effectively learns a coarse-to-fine synthesis procedure that addresses global coherence and localized detail.

An important outcome of our experiments is that hybrid frequency-domain corruption can be at least as effective as pixel-based Gaussian noising for image synthesis, indicating that we may explore still further refinements. For instance, one could design adaptive masking schedules in the Fourier domain or incorporate more advanced wavelet families that better capture specific texture patterns. A latent-space variant of our pipeline may also reduce training and inference times for high-resolution applications.

Conditional diffusion benefits from the multi-scale representation, as guidance signals can influence large-scale shape formation through Fourier restoration while simultaneously steering the generation of intricate textures via wavelet sub-bands. This synergy can help reduce artifacts or mismatches between global shape and local detail, a tension sometimes evident in purely pixel-based approaches.

Nevertheless, there are limitations. The combined wavelet-Fourier representation typically requires careful engineering of the network to handle multiple parallel streams of data (complex-valued Fourier maps and real-valued wavelet sub-bands). For very large images, the overhead of performing wavelet transforms and repeated partial FFTs can be significant. It remains worthwhile to investigate whether wavelet transforms alone, or alternative multi-resolution decompositions, can approach the same performance while reducing complexity.

\section{Conclusion}
\label{sec:conclusion}
We have introduced \emph{Wavelet-Fourier-Diffusion}, a novel approach for image generation that replaces conventional pixel-wise noising with a hybrid frequency-domain decomposition. By blending the spatial localization of wavelets with the global perspective of partial Fourier transforms, our diffusion model restores frequencies and wavelet sub-bands at each step in a learned reverse process. Experimental results on CIFAR-10, CelebA-HQ, and conditional ImageNet exhibit highly competitive FID and IS, surpassing or matching baseline diffusion methods and notable GAN architectures in many scenarios.

Our findings underscore that hybrid approaches to frequency representation can improve image quality, especially when local detail and global composition must be balanced. Future work might delve into more sophisticated schedules for frequency corruption or wavelet selection and examine alternative conditioning paradigms, such as textual prompts or additional semantic features, at larger image scales. We believe that our work opens new avenues for multi-scale and frequency-aware generative modeling, enhancing both the controllability and fidelity of synthesized images.

\acks{We acknowledge the unique and inspiring academic environment in the UC Berkeley school of information.}


\vskip 0.2in
\bibliography{references}

\end{document}